\documentclass[10pt,twocolumn,letterpaper]{article}

\usepackage{wacv}
\makeatletter
\@namedef{ver@everyshi.sty}{}
\makeatother
\usepackage{times}
\usepackage{epsfig}
\usepackage{latexsym}
\usepackage{microtype}
\usepackage{graphicx}
\usepackage{amsmath}
\usepackage{amssymb}
\usepackage{booktabs}
\usepackage{xfrac}
\usepackage{verbatim}
\usepackage{wrapfig}
\usepackage{multirow}
\usepackage{caption}
\usepackage{float}
\usepackage{subcaption}
\usepackage[switch]{lineno}
\usepackage[normalem]{ulem}
\useunder{\uline}{\ul}{}
\usepackage[accsupp]{axessibility}
\newcommand*\samethanks[1][\value{footnote}]{\footnotemark[#1]}
\usepackage{pdfpages} 
\usepackage{pgffor} 
\makeatletter
\AtBeginDocument{\let\LS@rot\@undefined}
\makeatother

\def\wacvPaperID{496} 

\wacvalgorithmstrack   

\wacvfinalcopy 

\ifwacvfinal
\usepackage[breaklinks=true,bookmarks=false]{hyperref}
\else
\usepackage[pagebackref=true,breaklinks=true,colorlinks,bookmarks=false]{hyperref}
\fi

\begin{document}

\title{Learning by Hallucinating: Vision-Language Pre-training with Weak Supervision}

\author{Tzu-Jui Julius Wang, Jorma Laaksonen\\
Aalto University, Finland\\
{\tt\small \{tzu-jui.wang, jorma.laaksonen\}@aalto.fi}
\and
Tomas Langer\\
Intuition Machines Inc.\\
{\tt\small tomas@intuitionmachines.com}
\and
Heikki Arponen\\
Systematic Alpha\thanks{Work done at Intuition Machines Inc.}\\
{\tt\small heikki.a.arponen@gmail.com}
\and
Tom E. Bishop\\
Glass Imaging\samethanks[1]\\
{\tt\small tom@glass-imaging.com}
}

\maketitle
\thispagestyle{empty}

\begin{abstract}
    Weakly-supervised vision-language (V-L) pre-training (W-VLP) aims at learning cross-modal alignment with little or no paired data, such as aligned images and captions.
    Recent W-VLP methods, which pair visual features with object tags, help achieve performances comparable with some VLP models trained with aligned pairs in various V-L downstream tasks. This, however, is not the case in cross-modal retrieval (XMR). We argue that the learning of such a W-VLP model is curbed and biased by the object tags of limited semantics.
    
    We address the lack of paired V-L data for model supervision with a novel \textbf{V}isual \textbf{V}ocabulary based \textbf{F}eature \textbf{H}allucinator (WFH), which is trained via weak supervision as a W-VLP model, not requiring images paired with captions. WFH generates visual hallucinations from texts, which are then paired with the originally unpaired texts, allowing more diverse interactions across modalities.
    
    Empirically, WFH consistently boosts the prior W-VLP works, e.g. U-VisualBERT (U-VB), over a variety of V-L tasks, i.e. XMR, Visual Question Answering, etc. Notably, benchmarked with recall@\{1,5,10\}, it consistently improves U-VB on image-to-text and text-to-image retrieval on two popular datasets Flickr30K and MSCOCO. Meanwhile, it gains by at least 14.5\% in cross-dataset generalization tests on these XMR tasks. Moreover, in other V-L downstream tasks considered, our WFH models are on par with models trained with paired V-L data, revealing the utility of unpaired data. These results demonstrate greater generalization of the proposed W-VLP model with WFH.
\end{abstract}
\vspace{-0.7cm}
\section{Introduction}
    \begin{figure}[h]
        \centering
        \begin{subfigure}[t]{.5\textwidth}
        \includegraphics[width=0.925\linewidth]{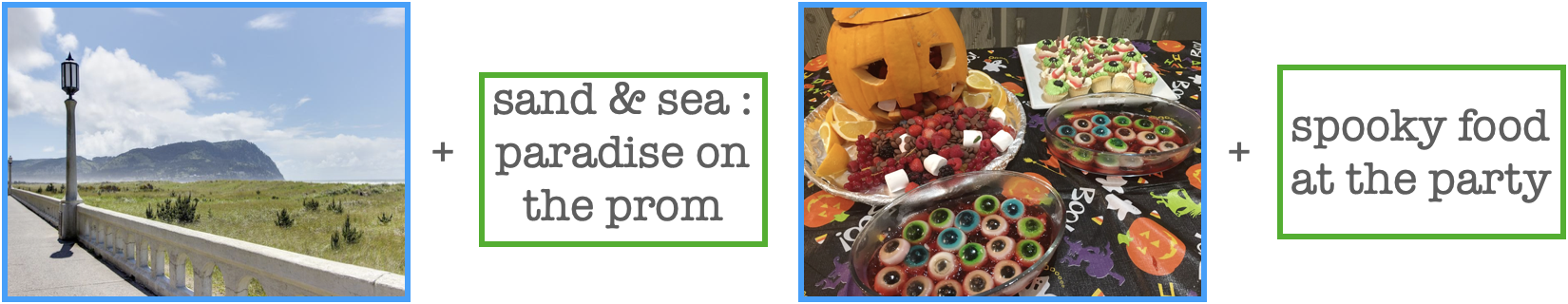}
        \caption{Fully-supervised pre-training}
        \label{fig:supervised}
        \end{subfigure}\\
        \begin{subfigure}[t]{.5\textwidth}
        \includegraphics[width=0.88\linewidth]{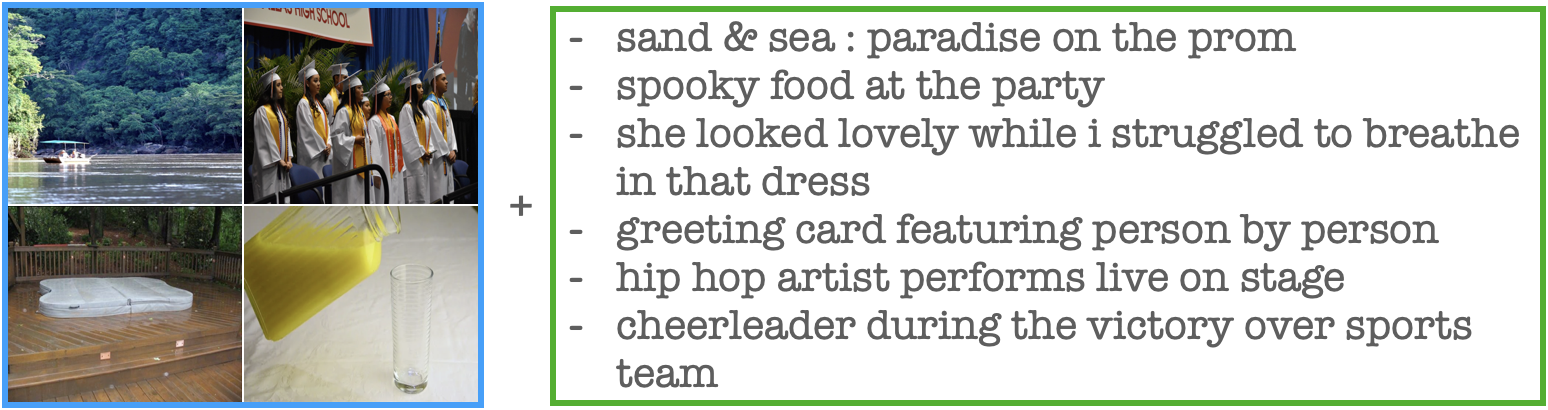}
        \caption{This work: Weakly-supervised pre-training}
        \label{fig:weakly}
        \end{subfigure}
        \caption{Examples of different pre-training settings: (a) The fully-supervised setting is given image-caption pairs, whereas this work focuses on (b) The weakly-supervised setting which learns on unpaired images and captions.}
        \label{fig:problem}
    \end{figure}
    \vspace{-3pt}
    Vision-language pre-training (VLP) has gained popularity as it shows great generalizability and transferability to many vision-language (V-L) downstream tasks. Pre-training is usually done on \textit{webly-supervised} datasets, which are collected semi-automatically through the Internet and are hence noisy, e.g. the image and captions can be of weak mutual relevance. Furthermore, these uncurated image-text pairs may contain a wide spectrum of inappropriate contents that lead to some daunting biases when taken to train a model \cite{birhane2021multimodal}. Despite trained on noisy datasets, these VLP models are shown to excel at various V-L downstream tasks \cite{alberti2019fusion,tan2019lxmert,lu2019vilbert,li2019visualbert,qi2020imagebert,li2020unicoder,su2019vl,chen2019uniter,li2020oscar,gan2020large,yu2020ernie,huo2021wenlan,zhang2021vinvl}. More recent works, such as CLIP \cite{radford2021learning} and ALIGN \cite{jia2021scaling}, enjoy greater downstream improvements being pre-trained on even larger amounts of image-text pairs. These excellent prior works, on the one hand, offer a promising direction -- a model properly pre-trained with massive amount of data, which could be imperfectly labeled, generalizes far better than one trained from scratch on a small dataset. On the other hand, the V-L research has been on a data-hungry path towards larger data collection efforts. This development could blur the other path more on trading-off the data efficiency and the generalization capability of V-L models. 
    
    \begin{figure}[ht]
        \centering
        \includegraphics[width=0.425\textwidth]{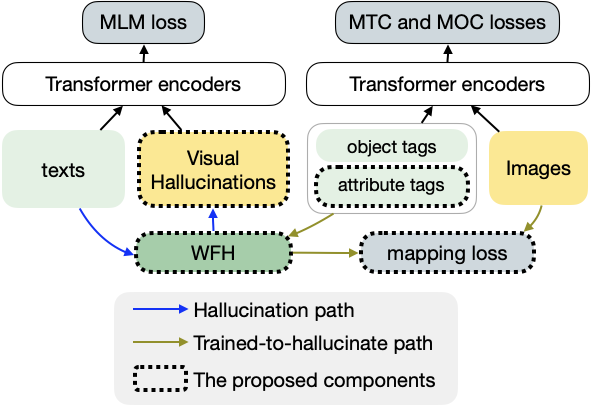}
        \caption{The proposed W-VLP model with the \textbf{V}isual \textbf{V}ocabulary based \textbf{F}eature \textbf{H}allucinator (WFH) at a glance. WFH is trained alongside to generate visual representations to pair with the textual counterparts. The components within the dotted frames distinguish us from the previous state-of-the-art W-VLP model, U-VisualBERT \cite{li-etal-2021-unsupervised}. Please refer to Sec.~\ref{sec:loss} for the losses and their abbreviations.}
        \label{fig:teaser}
        \vspace{-0.2cm}
    \end{figure}
    
    Two different perspectives to enhance the data efficiency have been suggested. The first adopts the self-knowledge distillation principle, which guides the learning with soft labels predicted by the exponentially-averaged self, i.e. the same model with the parameters being updated by the exponential moving average \cite{cheng2021data,grill2020bootstrap}. The second approach learns with limited access to paired images and texts \cite{hsu2018unsupervised,li-etal-2021-unsupervised}, thus largely reducing the effort in collecting a textual description for each image. This weakly-supervised setting makes VLP much more challenging since the aim of VLP is to learn to align V-L domains over paired data. Figure~\ref{fig:problem} illustrates the difference in the supervised and weakly-supervised settings.
    
    Weakly-supervised VLP (W-VLP), though being a crucial step to unleash the potential of abundant web images and texts, is much less explored than supervised VLP (S-VLP) and only explored in some specific domains, e.g. medical imaging \cite{dou2020unpaired}. Interestingly, we find that the recently proposed W-VLP models, e.g. the unsupervised VisualBERT (U-VB) \cite{li-etal-2021-unsupervised}, largely fall short on cross-modal retrieval (XMR) tasks, motivating us to improve a W-VLP model particularly on XMR tasks. Concretely, our work enhances one of the pioneering W-VLP works, i.e. U-VB, by capitalizing more on the pre-trained visual attribute and object detectors with a novel Visual Vocabulary based Feature Hallucinator (WFH). WFH, depicted in Figure~\ref{fig:teaser}, is trained similarly to a W-VLP model without directly training on massive amounts of paired data. The central idea of WFH is to generate visual counterparts from textual representations with layers of Transformer encoders. 
    The WFH-generated features are then paired with the originally unpaired texts. 
    
    It is worth clarifying that we do not claim the proposed model to be unsupervised (as is claimed for U-VB by its authors) but weakly-supervised. Both U-VB and our proposed model exploit knowledge from a pre-trained object detector for the follow-up unpaired training. Hence, they are exposed to some amounts of paired information, e.g. image regions and their object/attribute classes. We hereby consider that both models are learned under weak supervision.

    We summarize the contributions as follows: (1) We present a novel WFH that enables more interactions across modalities during pre-training. (2) We propose a W-VLP model that accommodates object tags, attribute tags and the WFH-generated features. (3) The proposed model consistently outperforms the state-of-the-art weakly-supervised baseline, U-VisualBERT (U-VB), on the XMR tasks (i.e. text-to-image, image-to-text retrieval, and cross-dataset generalization), Visual Question Answering (VQA), Referring Expression Comprehension (REC), and Visual Entailment (VE) tasks, on totally six datasets. (4) We provide studies on, e.g. expressiveness of the word token embeddings and behavior of the attention probabilities in the Transformer encoder, to better understand the inner working of the W-VLP models. The introduced WFH is simple but shown effective given these quantified results.
    
\section{Related Work}
We introduce related work starting from the advancements in S-VLP methods followed by the W-VLP methods. We then explore more applications, e.g. image translation \cite{zhu2017unpaired}, medical image segmentation \cite{valindria2018multi,dou2020unpaired}, unsupervised machine translation \cite{lample2018word} and unsupervised domain adaptation \cite{long2015learning,ghifary2016deep,tzeng2017adversarial,zhang2020label}, which advocate the usefulness of the unpaired data.
\subsection{Supervised V-L Pre-training}
    
    Most recently proposed VLP models adapt Transformer \cite{vaswani2017attention,devlin2019bert} for VLP with differences in architectures and training objectives. The VLP model architectures can be categorized into single- and two-stream models. The single-stream models, such as VisualBERT \cite{li2019visualbert}, ImageBERT \cite{qi2020imagebert}, Unicoder-VL \cite{li2020unicoder}, VL-BERT \cite{su2019vl}, UNITER \cite{chen2019uniter}, Oscar \cite{li2020oscar}, and SOHO \cite{huang2021seeing} etc., adopt a unified Transformer sharing the parameters across modalities. The two-stream models, e.g. LXMERT \cite{tan2019lxmert} and ViLBERT \cite{lu2019vilbert}, train a separate Transformer for each modality. These two separate Transformers cross-attend the representations from each layer of the other Transformer to learn cross-domain alignment through the attention mechanism. Though being architecturally simpler with less parameters to optimize, single-stream models are strongly comparable to two-stream models. 
    
    The usual training objectives of the VLP models are Masked Language Modeling (MLM) and Masked Region Modeling (MRM), with variants such as Masked Object Classification (MOC) and Masked Region Feature Regression (MRFR). Image-Text Alignment (ITA), which classifies if the V-L inputs are aligned, is used to learn V-L alignment on the sentence level. The optimal transport method \cite{cuturi2013sinkhorn} can be used to learn fine-grained alignment across image regions and words. Oscar \cite{li2020oscar} introduces object tags detected from the images as the anchors \cite{lample2018word} aligning word tokens and their visual groundings. VILLA \cite{gan2020large} improves other V-L frameworks by adding adversarial perturbation to the V-L input spaces. More recent works have been advancing VLP by, e.g. training with larger datasets \cite{radford2021learning,jia2021scaling} and enriching the image tags \cite{zhang2021vinvl}, which can benefit the framework such as Oscar. ALBEF \cite{li2021align} emphasizes cross-modal alignments in the early Transformer layers and learns from its momentum self to improve learning on noisy data.
    
    
\subsection{Weakly-supervised V-L Pre-training} 
    Aiming to pre-train a V-L model which learns to align V-L domains without image-text pairs, W-VLP is to save the substantial data collection effort. Hsu et al. \cite{hsu2018unsupervised} studied W-VLP in the context of medical imaging. Recently, Li et al. \cite{li-etal-2021-unsupervised} proposed U-VB to be trained without accessing the image-text pairs. It learns cross-domain alignment with object tags served as anchors between domains and considered as "fake" data paired with the images. However, U-VB's learning could be confined by those tags which only amount to 1,600 object classes from Visual Genome (VG) \cite{krishna2017visual} and bias the model to learn strong association between the visual and a limited amount of object tags' representations \cite{yang2021causal}.

    We thereby introduce a novel Visual Vocabulary based Feature Hallucinator (WFH), which aims to alleviate such a bias by generating regional visual representations to be paired with the textual description, e.g. a caption for an image. WFH generates diverse representations to offer a bridging signal across V-L domains. As a result, WFH greatly enhances U-VB over various V-L tasks.

\subsection{Applications in Learning from Unpaired Data}
    Research interest in learning from unpaired data has grown in various applications. Along with the great advancement in Generative Adversarial Networks (GANs) \cite{goodfellow2014generative}, learning to translate images from one domain to another with different styles or artistic touches has been shown feasible without paired images \cite{zhu2017unpaired}. Learning multi-modal representations for medical image analysis, e.g. organ segmentation, with unpaired CT and MRI scan images has also shown improvement in the segmentation accuracy compared to the models learned via a single modality \cite{valindria2018multi,dou2020unpaired}. Unsupervised machine translation \cite{lample2018word} and unsupervised domain adaptation \cite{long2015learning,ghifary2016deep,tzeng2017adversarial,zhang2020label} share similarity with W-VLP in that they both learn to transfer or align domains without having access to paired data.

\section{Our Proposed WFH Model}
    \begin{figure*}
        \centering
        \begin{subfigure}{.68\textwidth}
          \centering
          \includegraphics[width=\linewidth]{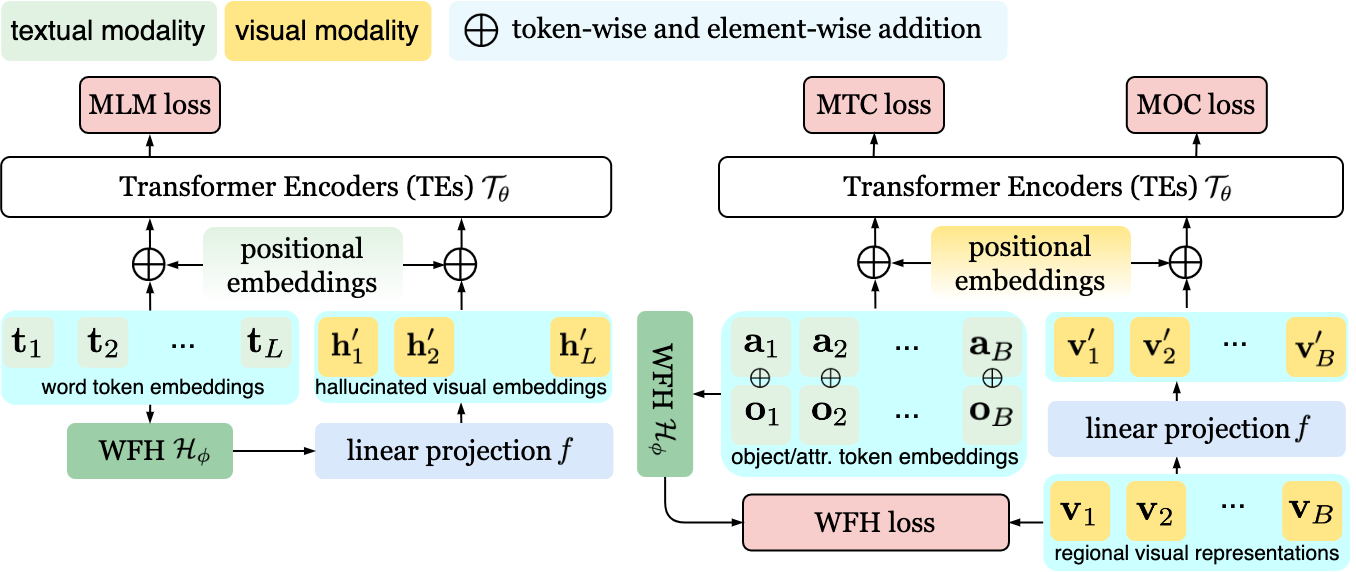}
          \caption{The proposed WFH model architecture and the training losses.}
          \label{fig:arch}
        \end{subfigure}\hfill%
        \begin{subfigure}{.30\textwidth}
          \centering
          \includegraphics[width=\linewidth]{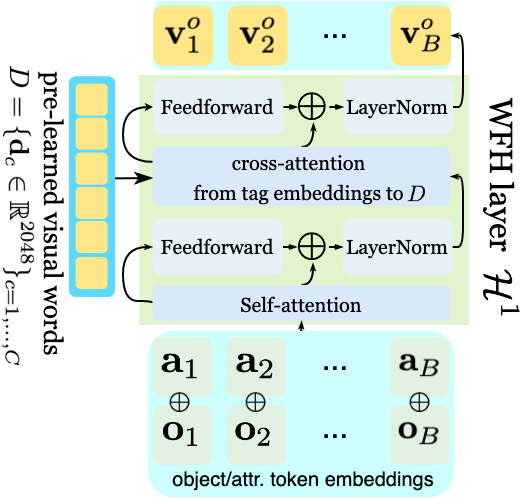}
          \caption{Illustration of a 1-layer WFH.}
          \label{fig:wfh}
        \end{subfigure}
        \caption{The proposed weakly-supervised V-L pre-training model architecture with the cross-domain Visual Vocabulary based Feature Hallucinator (WFH).}
        \label{fig:arch_wfh}
    \end{figure*}
The proposed W-VLP model with Visual Vocabulary based Feature Hallucinator (WFH), sketched in Figure~\ref{fig:arch}, consists of a single-stream Transformer $\mathcal{T}_{\theta}$ which takes multi-modal inputs and shares parameters, i.e. those associated with \textit{queries}, \textit{keys}, and \textit{values}, across modalities. Two sets of inputs are separately fed into $\mathcal{T}_{\theta}$. The first set $S_1 = \{(t_l, \textbf{h}_l)\}_{l=1}^L$ consists of $L$ text tokens $t_l$, each of which corresponds to a \textit{hallucinated} visual representation $\textbf{h}_l$, which we introduce in a later section. Another set of inputs $S_2 = \{(r_b, o_b, a_b)\}_{b=1}^B$ consists of (1) $B=36$ regions of interest $\{r_b\}_{b=1}^B$ generated from a pre-trained object detector $\mathcal{O}$, the predicted object class probabilities given by $\mathcal{O}$, and (2) the sampled object tag $o_b \sim P_b^{obj}$ and attribute tag $a_b \sim P_b^{attr}$, where $P_b^{obj}$ and $P_b^{attr}$ are the predicted probabilities over the object and attribute classes \footnote{We refer the object and attribute classes to VG's object and attribute classes. The examples of object classes are \textit{dishwasher}, \textit{cat}, \textit{ocean}, etc; attributes are \textit{blank}, \textit{matallic}, \textit{talking}, etc.} obtained from $\mathcal{O}$, respectively. $\mathcal{T}_{\theta}$ adopts the same Transformer architecture as in U-VB.


\subsection{Model Architecture}
This section focuses on formulating V-L inputs, WFH, the pre-training objectives and the losses. The differences with U-VB are emphasized.
\vspace{-0.25cm}
\subsubsection{V-L Inputs from $S_1$ set} 
    Each language token $t_l$ from the token sequence $\{t_l\}_{l=1}^L$ is obtained by tokenizing a sentence and embedded as 
        \begin{equation}
            \textbf{t}_l = T(T_{BERT}(t_l)) \in \mathbb{R}^{768}, l=1,...,L,
        \end{equation}
    where $T_{BERT}(\cdot)$ is the BERT's embedding and $T$ is a linear embedding layer. Each hallucinated visual representation is generated from the proposed WFH $\mathcal{H}_{\phi}$, i.e.
        \begin{align}
            \textbf{h}_l &= \mathcal{H}_{\phi}(t_l | \{t_i\}_{i=1}^L, D) \in \mathbb{R}^{2048}, \, l=1,...,L, \\
            \textbf{h}'_l &= f(\textbf{h}_l) = W_f \textbf{h}_l + b_f \in \mathbb{R}^{768},\, l=1,...,L,
        \end{align}
    where $D=\{\textbf{d}_c \in \mathbb{R}^{2048}\}_{c=1}^C$ is the pre-learned visual dictionary. $\mathcal{H}_{\phi}(\cdot)$ is the hallucinator function which we will formally introduce later in Eqs.~(\ref{eq:wfh_v}) and (\ref{eq:wfh_h}). $f$ is a linear projection function parameterized by learnable weights $W_f \in \mathbb{R}^{768 \times 2048}$ and biases $b_f \in \mathbb{R}^{768}$.  $\textbf{t}_l$ and $\textbf{h}'_l$ are respectively added with the token positional embedding $\textbf{p}_l^{W} \in \mathbb{R}^{768}$ obtained by linearly transforming $l \in [1,...,L]$, which is the token's position in the sequence. $\phi$ denotes WFH's parameters.
\vspace{-0.25cm}
\subsubsection{V-L Inputs from $S_2$ set}
    We denote the regional visual representations as $\{\textbf{v}_b \in \mathbb{R}^{2048}\}_{b=1}^B$. Each $\textbf{v}_b$ is extracted from $r_b$ by $\mathcal{O}$ and transformed to $\textbf{v}'_b$ via
        \begin{equation}\label{eq:v_trans}
            \textbf{v}'_b = f(\textbf{v}_b) \in \mathbb{R}^{768},
        \end{equation}
    which is consistent with the size of BERT's token embeddings.
    $o_b$ and $a_b$ are pre-processed similarly: tokenized by BERT's WordPiece tokenizer \cite{devlin2019bert} and transformed to
        \begin{align}
            \textbf{o}_b &= T(T_{BERT}(o'_b)) \in \mathbb{R}^{768}, b=1,...,B,\label{eq:token_transform}\\
            \textbf{a}_b &= T(T_{BERT}(a'_b)) \in \mathbb{R}^{768}, b=1,...,B,\label{eq:token_transform2}
        \end{align}
    respectively. $o'_b$ and $a'_b$ are the tokenized tags\footnote{Note that each tag can be tokenized to more than one tokens with the WordPiece tokenizer. We keep the same subscript $b$ as in $r_b, o_b, a_b$ for notational simplicity.}. For the visual inputs, $\textbf{v}'_b$ is added with the image positional embedding, $\textbf{p}_b^{I} \in \mathbb{R}^{768}$, linearly transformed from a vector of the normalized bounding box x- and y-coordinates, width, and height of $r_b$. For the language inputs, we have $\textbf{o}_b + \textbf{a}_b + \textbf{p}_b^I$. We fuse $\textbf{o}_b$ and $\textbf{a}_b$ embeddings by summing while other ways, e.g. appending the tokens, increases the number of input tokens, incurring much higher training complexity to the Transformer, which is quadratic to the number of tokens. 

\subsubsection{Differences with U-VB} 
    Figure~\ref{fig:teaser} highlights the differences between the proposed WFH model and U-VB. First, the WFH model additionally augments each object tag embedding with its attribute embedding. Second, U-VB processes $t_l$ alone without a visual counterparts of any kind, unlike us pairing $t_l$ with its hallucinations; in other words, the language part of U-VB's training is just fine-tuning BERT's weights on the given texts.

\subsection{Learning Visual Hallucinator}\label{subsec:wfh}
    As shown in Figure~\ref{fig:wfh}, A WFH layer takes a textual token whose visual counterpart is to be hallucinated. The token is processed to account for its context in the sequence with a self-attention mechanism. The self-attention outputs then attend across modalities to each pre-learned $\textbf{d}_c \in D$ to hallucinate visual representations. One can stack more WFH layers to model more complex interactions. The visual dictionary $D$ is learned off-line by simple K-means with momentum updates \cite{huang2021seeing} on the regional visual representations extracted from Conceptual Captions (CC) \cite{sharma2018conceptual} images. Please find how $D$ is learned in detail in the supplementary material (SM). Formally, the input to WFH can be a sequence of textual tokens $\{\textbf{t}_l\}_{l=1}^L$ or $\{ \textbf{o}'_b \}_{b=1}^B$ with each $\textbf{o}'_b$ obtained via
    \begin{equation}
        \textbf{o}'_b = \textbf{o}_b + \textbf{a}_b, \, b=1,...,B.
    \end{equation}
    When given $\{\textbf{t}_l\}_{l=1}^L$, WFH generates visual representations $\{\textbf{v}^{t}_l\}_{l=1}^L$. When given $\{ \textbf{o}'_b \}_{b=1}^B$, WFH generates $\{\textbf{v}^{o}_b\}_{b=1}^B$. For example, generating $\textbf{v}^{o}_b$ (as illustrated in Figure~\ref{fig:wfh}) can be formulated as
        \begin{align}
            \textbf{v}^{o}_b &= \mathcal{H}_{\phi}(\textbf{o}'_b, \{ \textbf{o}'_i \}_{i=1,i \neq b}^B, D) \in \mathbb{R}^{2048}, \label{eq:wfh_v}\\
            \mathcal{H}_{\phi} &= \mathcal{H}^J \circ \mathcal{H}^{J-1} \circ \dotsm \circ \mathcal{H}^{1}, \label{eq:wfh_h}
        \end{align}
    where $\circ$ is the function composition. For $j=1,...,J,$
        \begin{align}
            \mathcal{H}^j (\cdot,\cdot,D) &= \big\Vert_{m=1}^M \{A_x^{j,m}(Q^{j,m}, K^{j,m}, V^{j,m})\}, \label{eq:xatt}\\
            Q^{j,m} &= \{W^{j,m}_Q \textbf{o}{''}_i^j\}_{i=1}^B, \\
            \textbf{o}{''}_b^j &= A_s^j(\textbf{o}{'}_b^j | \{\textbf{o}{'}_i^j\}_{i=1}^B) \in \mathbb{R}^{768}, \, \textbf{o}{'}_b^1 = \textbf{o}'_b, \, \\
            K^{j,m} &= \{W^{j,m}_K \textbf{d}_c\}_{c=1}^C, \\
            V^{j,m} &= \{W^{j,m}_V \textbf{d}_c\}_{c=1}^C.
        \end{align}
    $\Vert$ indicates the concatenation of vectors, $J$ is the number of WFH layers and in each layer $A_s^j$ is the \textit{self-attention} and $A_x^{j,m}$ is the $m^{th}$ attention head (of totally $M$ heads) in the \textit{cross-attention} layer.

\subsubsection{Self-attention $A_s^j$} 
    $A_s^j$ produces contextual textual representations $\textbf{o}{''}_i^j$, which are input to each $A_x^{j,m}$ to construct the \textit{query}. $A_s^j$ is the multi-head attention mechanism introduced in \cite{vaswani2017attention} with $M=12$ heads. Each head in $A_s^j$ produces a $\sfrac{768}{12}=64\text{-dimensional}$ vector, and $\textbf{o}{''}_b^j$ is the concatenation of those 12 vectors.
\vspace{-0.25cm}
\subsubsection{Cross-attention $A_x^{j,m}$} 
    The textual \textit{queries} in $Q^j$ learn to align with the visual \textit{keys} in $K^{j,m}$ and generate visual representations by $\mathcal{H}^j$. $W^{j,m}_Q,\,W^{j,m}_K,\,W^{j,m}_V$ are the learnable weight matrices for \textit{queries}, \textit{keys}, and \textit{values}, respectively. $W_Q^{j,m} \in \mathbb{R}^{\sfrac{768}{M} \times 768}$, $W_K^{j,m} \in \mathbb{R}^{\sfrac{768}{M} \times 2048}$, and $W_V^{j,m} \in \mathbb{R}^{\sfrac{2048}{M} \times 2048}$ with $M=16$ for $j=J$, i.e. in the last layer; otherwise, $W_V^{j,m} \in \mathbb{R}^{\sfrac{768}{M} \times 2048}$ with $M=12$. Concatenating $M$ vectors of each head produces the output of a WFH layer. One generates $\textbf{v}^{t}_i$ given $\{t_l\}_{l=1}^L$ with the same process. 
\vspace{-0.25cm}
\subsubsection{WFH Objectives} 
    WFH is learned (1) implicitly through achieving Masked Language Modeling (MLM) task \cite{li-etal-2021-unsupervised} with the captions, and (2) explicitly with the mapping loss:
    \begin{equation}\label{eq:wfh}
        L_{\phi,\text{WFH}} = \frac{1}{B} \sum_{i=1}^B ||\textbf{v}_i^o - \textbf{v}_i||_2^2, 
    \end{equation}
    where we regress $\textbf{v}_i^o$, hallucinated from the object/attribute tag embedding $\textbf{o}'_i$, to $\textbf{v}_i \in \mathbb{R}^{2048}$, the visual representation extracted from $\mathcal{O}$. This objective ensures that the hallucinated features stay close to the visual domain in which the real $\textbf{v}_i$ features reside.
\vspace{-0.25cm}    
\subsubsection{Design Consideration for WFH}
    An alternative to Eq.~(\ref{eq:wfh}) is to learn a direct projection matrix for mapping the textual representations to the visual ones. However, this is challenging in practice as it involves transforming high-dimensional distributions from one modality to another. Instead, the WFH's hallucination process is only to retrieve a visual representation from the space constructed by $D$, hence it avoids the direct mapping across domains and produces better hallucinations.

\subsubsection{Visualizing Hallucinations}    
    Please refer to the SM for the visualization, which shows that the hallucinated features (1) are contextual, and (2) appear to serve as the bridging representations across the V-L domains.
    

\subsection{Loss Functions}\label{sec:loss}
$\mathcal{T}_{\theta}$ and WFH $\mathcal{H}_{\phi}$ are trained with the overall loss $L_{\theta, \phi}$ for an unpaired image and caption:
    \begin{equation}\label{eq:total_loss}
        L_{\theta, \phi} = L_{\theta,\text{MLM}} + L_{\theta,\text{MTC}} + L_{\theta,\text{MOC}} + L_{\phi,\text{WFH}}, 
    \end{equation}
which is composed of equally weighted Masked Language Modeling (MLM), Masked Tag Classification (MTC), Masked Object Classification (MOC) and WFH losses. MLM is to predict the masked tokens in the sequence $\{t_l\}_{l=1}^L$. Our MLM is conditioned on the hallucinated $\textbf{v}^{t}_i$ along with the word tokens given at the input. MTC is to predict the masked object tag tokens. MOC is to predict the object class for the masked, i.e. zeroed-out, visual features. We closely follow U-VB's MTC and MOC including, e.g. tokens are masked at 15\% of probability. The attribute tokens are always not masked. $L_{\phi,\text{WFH}}$ is the proposed WFH's objective given in Eq.~(\ref{eq:wfh}). It is worth noting that WFH's learning is not only explicitly guided by $L_{\phi,\text{WFH}}$, but also implicitly by $L_{\theta,\text{MLM}}$ to generate useful features that also benefit the MLM task.

\subsection{V-L Downstream Tasks}\label{sec:vl_downstream}
\subsubsection{XMR Tasks} Following the same methodology as other VLP works, we fine-tune $\mathcal{T}_{\theta}$ with two additional projection layers for XMR. Specifically, $\mathcal{T}_{\theta}$ outputs contextual representations $\{\textbf{t}'_l\}_{l=1}^L$ for the caption and $\{\textbf{v}''_b\}_{b=1}^B$ for the image given a paired image and caption. We predict the matching score $s$ by
    \begin{align}
        s &= \gamma \cdot \cos( f_t(\bar{\textbf{t}}),
                                f_v(\bar{\textbf{v}})), \label{eq:cos}\\
        \bar{\textbf{t}} &= \frac{1}{L} \sum_{l=1}^L \textbf{t}'_l, \, \bar{\textbf{v}} = \frac{1}{B} \sum_{b=1}^B \textbf{v}''_b. \label{eq:avg_pool}
    \end{align}
$f_t(\cdot)$ and $f_v(\cdot)$ are linear projections which do not change the dimensionalities of their inputs. Note that we summarize the image and caption with mean-pooled token embeddings instead of a single token embedding from, e.g. the class token $\texttt{[CLS]}$ as in other VLP works \cite{lu2019vilbert}. We find that using mean-pooled embeddings leads to slightly better performance, aligning the finding in \cite{ma2019universal}. As in \cite{lu2019vilbert}, the training objective is a 4-way classification that involves selecting three distracting choices for each image-text pair.
\subsubsection{VQA, VE, and REC Tasks} 
For VQA, the model predicts the distribution over $N_a$ answers on $\textbf{c} \in \mathbb{R}^{1536}$, 
    \begin{equation}\label{eq:concat_c}
        \textbf{c} = \bar{\textbf{t}} \, || \, \bar{\textbf{v}},,
    \end{equation}
i.e. the concatenation of $\bar{\textbf{t}}$ and $\bar{\textbf{v}}$, which is fed to a linear projection layer whose width is 1,024, followed by a GeLU \cite{hendrycks2016gaussian} activation function and a $N_a$-way classification layer. Similarly for VE, the model predicts the answer distribution via passing $c$ to a linear layer. For REC, to predict the visual grounding score for each image region, the model feeds $\textbf{v}''_b$ (from Eq.~(\ref{eq:avg_pool}), $b=1, ..., B$) to a linear layer that outputs 768 neurons, which are then processed by GeLU and another linear layer producing the final matching score.

\section{Experiments}
    This section introduces datasets for pre-training and fine-tuning for various V-L tasks. We detail the experiment settings followed by comparisons with other W-VLP methods. 
\subsection{Datasets and Tasks}
    \subsubsection{Pre-training Dataset} The object detector that generates object and attribute tags is trained on VG. The visual vocabulary devised in WFH is fixed and pre-learned on the regional representations extracted from CC images with the same object detector. All the W-VLP models throughout the experiments are pre-trained on CC by randomly selecting a batch of captions and images which are not paired. Particularly, we use 2.7M images and captions from CC.

    \subsubsection{XMR Datasets and Tasks} The pre-trained models are fine-tuned on image-caption pairs from MSCOCO \cite{lin2014microsoft} or Flickr30K \cite{plummer2015flickr30k} to study their transferability. For Flickr30K, we follow the training/validation/test splits as in \cite{karpathy2015deep}. For MSCOCO, we follow the splits as in \cite{lee2018stacked,qi2020imagebert,li2020unicoder}, and report the numbers by averaging over five folds of the test set, i.e. the "COCO 1K test set". We consider the following tasks: (1) image-to-Text Retrieval (TR), (2) text-to-Image Retrieval (IR), and (3) cross-dataset TR and IR, i.e. fine-tuned on COCO and tested on Flickr30K and vice versa.
    
    \subsubsection{VQA, REC, and VE Datasets} We assess the VQA task on two popular datasets: VQAv2 \cite{antol2015vqa,balanced_vqa_v2} and GQA \cite{hudson2019gqa}. VE and REC are evaluated on SNLI-VE \cite{xie2019visual} and RefCOCO+ \cite{yu2016modeling} datasets, respectively.   
\subsection{Model Parameters and Training Details}
    We develop our projects upon VOLTA
    \cite{bugliarello2020multimodal}, which is built with PyTorch \cite{NEURIPS2019_9015} and aims for speeding up multi-modal machine learning research by establishing baselines within a controlled setup, e.g. models trained on same amount of text-image pairs across different VLP models. The object detector uses ResNet-101 \cite{he2016deep} as the backbone. We follow the U-VB architecture, where each Transformer layer has $M=12$ attention heads and the dimensionality of the hidden state is 768. The size of the pre-learned visual dictionary $C$ is chosen from $\{1024, 1536, 3072\}$; the number of WFH layers $J$ from $\{1, 2, 3\}$; $\gamma$ in Eq.~(\ref{eq:cos}) from $\{8, 16, 32\}$. Throughout the experiments, the methods annotated with WFH are always trained with the attribute tokens added unless otherwise specified.

    All W-VLP models are trained with eight 16GB-V100 GPUs with batch size of 400 for 12 epochs. AdamW \cite{loshchilov2018decoupled} is used as the optimizer with weight decay as 0.01. The learning rate is adjusted with the warmup period being 10\% of the total epochs. It is peaked at $1.5625 \times 10^{-4}$ and linearly reduced to 0. The pre-training takes roughly one day for each  model. At the fine-tuning stage, the models are trained with batch sizes of 64, 256, 256, and 192 for XMR, VQA, REC, and VE, respectively, with two 16GB-V100 GPUs. AdamW is used with weight decay being 0.0001.

\begin{table*}[ht]
\small
\centering
\caption{Comparing models on Flickr30K. We borrow results of U-VisualBERT (U-VB) from \cite{li-etal-2021-unsupervised}, in which it is trained on 3M images and 5.5M captions (CC + BookCorpus). The models with $\dagger$ and our WFH models are replicated and implemented with VOLTA, and all of which are trained only on 2.7M CC images and captions for fair comparison. From one row after U-VB $\dagger$, it shows results of our proposed WFH models of different configurations, i.e. without attribute tokens involved (-attr) and different numbers of WFH layers employed, with the dictionary size $C=1024$. The best models are highlighted in bold and the second best is underlined. Meta Sum is the sum of R@1, 5, and 10. The results shown with $\pm$ are the means and the standard deviations obtained with five pre-training runs with different random seeds.}
\scalebox{0.75}{\setlength\tabcolsep{6pt}
\begin{tabular}{|l|llll|llll|}
\hline
\multirow{2}{*}{Models}                                & \multicolumn{4}{l|}{Text-Image Retrieval}                                                                                     & \multicolumn{4}{l|}{Image-Text retrieval}                                                                                     \\ \cline{2-9} 
                                                       & \multicolumn{1}{l|}{R@1}           & \multicolumn{1}{l|}{R@5}           & \multicolumn{1}{l|}{R@10}          & Meta Sum       & \multicolumn{1}{l|}{R@1}           & \multicolumn{1}{l|}{R@5}           & \multicolumn{1}{l|}{R@10}          & Meta Sum       \\ \hline
SCAN \cite{lee2018stacked}            & \multicolumn{1}{l|}{48.6}          & \multicolumn{1}{l|}{77.7}          & \multicolumn{1}{l|}{85.2}          & 211.5          & \multicolumn{1}{l|}{67.4}          & \multicolumn{1}{l|}{90.3}          & \multicolumn{1}{l|}{95.8}          & 253.5          \\ \hline
SCG \cite{shi2019knowledge}           & \multicolumn{1}{l|}{49.3}          & \multicolumn{1}{l|}{76.4}          & \multicolumn{1}{l|}{85.6}          & 211.3          & \multicolumn{1}{l|}{71.8}          & \multicolumn{1}{l|}{90.8}          & \multicolumn{1}{l|}{94.8}          & 257.4          \\ \hline
PFAN \cite{wang2019position}          & \multicolumn{1}{l|}{50.4}          & \multicolumn{1}{l|}{78.7}          & \multicolumn{1}{l|}{86.1}          & 215.2          & \multicolumn{1}{l|}{70.0}          & \multicolumn{1}{l|}{91.8}          & \multicolumn{1}{l|}{95.0}          & 256.8          \\ \hline
GPO \cite{chen2021learning}     & \multicolumn{1}{l|}{60.8}          & \multicolumn{1}{l|}{86.3}          & \multicolumn{1}{l|}{92.3}          & 239.4          & \multicolumn{1}{l|}{80.7}          & \multicolumn{1}{l|}{96.4}          & \multicolumn{1}{l|}{98.3}          & 275.4          \\ \hline\hline
UNITER \cite{chen2019uniter} $\dagger$                                       & \multicolumn{1}{l|}{62.2}          & \multicolumn{1}{l|}{85.9}          & \multicolumn{1}{l|}{91.6}          & 239.7          & \multicolumn{1}{l|}{77.8}          & \multicolumn{1}{l|}{92.2}          & \multicolumn{1}{l|}{96.0}          & 266.0            \\ \hline\hline
U-VB \cite{li-etal-2021-unsupervised} & \multicolumn{1}{l|}{55.4}          & \multicolumn{1}{l|}{82.9}          & \multicolumn{1}{l|}{89.8}          & 228.3          & \multicolumn{1}{c|}{-}             & \multicolumn{1}{c|}{-}             & \multicolumn{1}{c|}{-}             &                \\ \hline
U-VB $\dagger$                                         & \multicolumn{1}{l|}{54.4$\pm$0.3}          & \multicolumn{1}{l|}{81.7$\pm$0.4}          & \multicolumn{1}{l|}{88.8$\pm$0.3}          & 224.2$\pm$1.1          & \multicolumn{1}{l|}{67.8$\pm$0.3}          & \multicolumn{1}{l|}{90.7$\pm$0.5}    & \multicolumn{1}{l|}{\underline{94.9}$\pm$0.8} & 253.5$\pm$1.1    \\ \hline
U-VB (+attr) $\dagger$                  & \multicolumn{1}{l|}{52.5}          & \multicolumn{1}{l|}{81.3}          & \multicolumn{1}{l|}{88.3}          & 222.1          & \multicolumn{1}{l|}{65.5}          & \multicolumn{1}{l|}{89.7}    & \multicolumn{1}{l|}{94.8} & 250    \\ \hline
Ours: 1-layer WFH (-attr)                    & \multicolumn{1}{l|}{54.6}          & \multicolumn{1}{l|}{\underline{82.9}} & \multicolumn{1}{l|}{89.0}          & 226.5          & \multicolumn{1}{l|}{69.9}          & \multicolumn{1}{l|}{90.0}          & \multicolumn{1}{l|}{94.3}          & 254.2          \\ \hline
Ours: 1-layer WFH                           & \multicolumn{1}{l|}{{\ul 55.0}}    & \multicolumn{1}{l|}{82.7}          & \multicolumn{1}{l|}{\underline{89.8} } & {\ul 227.5}    & \multicolumn{1}{l|}{71.7}          & \multicolumn{1}{l|}{\textbf{91.4}}          & \multicolumn{1}{l|}{94.8}          & {\ul 257.9}          \\ \hline
Ours: 2-layer WFH                           & \multicolumn{1}{l|}{\textbf{56.4}$\pm$0.3 } & \multicolumn{1}{l|}{\textbf{83.2}$\pm$0.7}    & \multicolumn{1}{l|}{\textbf{89.9}$\pm$0.3}    & \textbf{229.5}$\pm$0.9 & \multicolumn{1}{l|}{\textbf{72.0}$\pm$0.4} & \multicolumn{1}{l|}{\underline{91.3}$\pm$0.5} & \multicolumn{1}{l|}{\textbf{95.6}$\pm$0.7}    & \textbf{258.9}$\pm$1.0 \\ \hline
\end{tabular}}
\label{table:stoa_flickr}
\end{table*}

\subsection{Quantitative Results}
    In what follows, we present results of each task considered and provide studies on (1) the effects brought by varying variables in the proposed WFH, (2) spectral analysis \cite{wang2019improving} on the text token matrices, and (3) different patterns of the attention probabilities over the Transformer layers from attention heads learned by models with and without WFH. We aim for providing better understanding of how our and U-VB models perform differently in the latter two studies.
\subsubsection{XMR Tasks}
    The main results of models on Flickr30K and MSCOCO are shown in Tables~\ref{table:stoa_flickr} and \ref{table:stoa_coco}, respectively. UNITER's V-L paired results serve as an upper bound for the W-VLP models. Results from SCAN \cite{lee2018stacked}, SCG \cite{shi2019knowledge}, PFAN \cite{wang2019position}, and GPO \cite{chen2021learning} are listed only for reference as they are task-specific and not proposed as generic VLP models. The models with $\dagger$ are replicated with VOLTA. Our work and U-VB share the same architecture, which adds a few additional task-specific layers on top of the pre-trained Transformer layers.

    On Flickr30K, we first show the recalls of our model trained with 1-layer WFH without attribute tokens. While most recall values are comparable with U-VB, we observe a clear gain in R@1 on TR. Adding attribute tokens improves R@\{1,5\} on TR and all recall values on IR over U-VB. We obtain the best results with 2-layer WFH with 3.7\% and 6.2\% gains in R@1 on IR and TR, respectively. We take this model to continue the comparisons with other models in the rest of the experiments. On MSCOCO, the proposed model consistently surpasses U-VB in every recall value. 

\begin{table}[h!]
\small
\centering
\caption{Comparing models on MSCOCO, 1K test set. We replicate U-VB (U-VB $\dagger$) as U-VB's authors did not report results on MSCOCO.} 
\vspace{-5pt}
\scalebox{0.75}{\setlength\tabcolsep{6pt}
\begin{tabular}{|l|l|l|l|l|l|l|}
\hline
\multirow{2}{*}{Models}                 & \multicolumn{3}{l|}{Text-Image Retrieval}                                       & \multicolumn{3}{l|}{Image-Text Retrieval}                                       \\ \cline{2-7} 
                                  & \multicolumn{1}{l|}{R@1} & \multicolumn{1}{l|}{R@5} & \multicolumn{1}{l|}{R@10} & \multicolumn{1}{l|}{R@1} & \multicolumn{1}{l|}{R@5} & \multicolumn{1}{l|}{R@10} \\ \hline
SCAN                              & 58.8                    & 88.4                    & 94.8                     & 72.7                    & 94.8                    & 98.4                     \\ \hline
SCG                               & 61.4                    & 88.9                    & 95.1                     & 76.6                    & 96.3                    & 99.2                     \\ \hline
PFAN                              & 61.6                    & 89.6                    & 95.2                     & 76.5                    & 96.3                    & 99.0                     \\ \hline
GPO                             & 64.8                    & 91.6                    & 96.5                     & 80.0                    & 97.0                    & 99.0                     \\ \hline\hline
U-VB $\dagger$ & 59.0$\pm$0.4                    & 88.0$\pm$0.2                    & 94.4$\pm$0.2                     & 73.0$\pm$0.4                    & 93.4$\pm$0.3                    & 97.3$\pm$0.3                     \\ \hline
WFH & \textbf{61.9}$\pm$0.6           & \textbf{89.4}$\pm$0.5                    & \textbf{95.3}$\pm$0.1            & \textbf{73.9}$\pm$0.1                    & \textbf{94.6}$\pm$0.3                    & \textbf{98.0}$\pm$0.4                     \\ \hline
\end{tabular}}
\label{table:stoa_coco}
\end{table}
\vspace{-10pt}
\subsubsection{Cross-dataset Generalization on XMR}
    When trained and tested with different datasets, both models suffer from perceivable drops in recall values compared to when trained and tested with the same dataset. Nevertheless, the proposed model significantly outperforms U-VB as shown in Table~\ref{table:cross}. Notably, the improvements in recall over U-VB are always higher when both models are trained on Flickr30K than on MSCOCO, i.e. 14.5\% and 16.5\% gains in R@1 compared to 3.93\% and 7.33\% on IR and TR, respectively. This indicates that the proposed model could generalize better to a smaller dataset, e.g. Flickr30K (29K training images), which is about three times smaller than MSCOCO (82K training images). 
    \begin{table}[h!]
    \small
    \centering
    \caption{Comparing models on cross-dataset generalization. }
    \scalebox{0.75}{\setlength\tabcolsep{6pt}
    \begin{tabular}{|l|l|l|l|l|l|l|}
    \hline
    \multirow{3}{*}{Models} & \multicolumn{3}{l|}{Text-Image Retrieval}                                                  & \multicolumn{3}{l|}{Image-Text Retrieval}                                       \\ \cline{2-7} 
                            & \multicolumn{6}{l|}{Flickr30K train - MSCOCO test}                                                                                                                           \\ \cline{2-7} 
                            & R@1                                 & \multicolumn{1}{l|}{R@5} & \multicolumn{1}{l|}{R@10} & \multicolumn{1}{l|}{R@1} & \multicolumn{1}{l|}{R@5} & \multicolumn{1}{l|}{R@10} \\ \hline
    U-VB $\dagger$                    & \multicolumn{1}{r|}{37.0}          & 69.3                    & 80.9                     & 45.6                    & 73.0                    & 83.3                     \\ \hline
    WFH                     & \multicolumn{1}{r|}{\textbf{42.3}} & \textbf{73.3}           & \textbf{84.3}            & \textbf{53.1}           & \textbf{79.2}           & \textbf{87.2}            \\ \hline
    \multirow{2}{*}{}       & \multicolumn{6}{l|}{MSCOCO train - Flickr30K test}                                                                                                                           \\ \cline{2-7} 
                            & R@1                                 & \multicolumn{1}{l|}{R@5} & \multicolumn{1}{l|}{R@10} & \multicolumn{1}{l|}{R@1} & \multicolumn{1}{l|}{R@5} & \multicolumn{1}{l|}{R@10} \\ \hline
    U-VB $\dagger$                   & \multicolumn{1}{r|}{45.2}          & 72.1                    & 81.0                     & 54.6                    & 80.0                    & 87.9                     \\ \hline
    WFH                     & \multicolumn{1}{r|}{\textbf{47.0}} & \textbf{73.8}           & \textbf{82.4}            & \textbf{58.6}           & \textbf{83.9}           & \textbf{90.5}            \\ \hline
    \end{tabular}}
    \label{table:cross}
    \end{table}
    \begin{table*}[]
    \small
    \caption{Comparing models fine-tuned for the VQA, REC and VE tasks. SOTA refers to the state-of-the-art task-specific models specified in Sec.~\ref{sec:vqa_rec_ve_results}. Any model with $\dagger$ refers to the replications realized in the VOLTA framework. VB w/o pt $\dagger$ represents the VisualBERT baseline without pre-training on image-text pairs. WFH refers to the same model as in Tables~\ref{table:stoa_coco} and \ref{table:cross}. The better models between U-VB $\dagger$ and WFH are highlighted in bold.}
    \centering
    \scalebox{0.75}{\setlength\tabcolsep{6pt}
    \begin{tabular}{|l|llll|l|lll|l|}
    \hline
                & \multicolumn{4}{l|}{VQAv2 test-dev}                                                              & GQA      & \multicolumn{3}{l|}{RefCOCO+}                                      & SNLI-VE \\ \hline
                & \multicolumn{1}{l|}{overall} & \multicolumn{1}{l|}{yes/no} & \multicolumn{1}{l|}{number} & other & test-dev & \multicolumn{1}{l|}{test set} & \multicolumn{1}{l|}{testA} & testB & test    \\ \hline
    SOTA        & \multicolumn{1}{l|}{70.63}   & \multicolumn{1}{l|}{86.82}  & \multicolumn{1}{l|}{53.26}  & 60.72 & 63.17    & \multicolumn{1}{l|}{-}        & \multicolumn{1}{l|}{75.13} & 66.17 & 71.16   \\ \hline
    VB w/o pt $\dagger$  & \multicolumn{1}{l|}{66.07}   & \multicolumn{1}{l|}{82.74}  & \multicolumn{1}{l|}{46.51}  & 56.29 & 53.55    & \multicolumn{1}{l|}{67.81}    & \multicolumn{1}{l|}{75.41} & 58.91 & 74.56   \\ \hline
    VB $\dagger$     & \multicolumn{1}{l|}{68.20}   & \multicolumn{1}{l|}{-}      & \multicolumn{1}{l|}{-}      & -     & 56.58    & \multicolumn{1}{l|}{69.70}    & \multicolumn{1}{l|}{-}     & -     & 75.67   \\ \hline
    U-VB \cite{li-etal-2021-unsupervised} & \multicolumn{1}{l|}{70.74}   & \multicolumn{1}{l|}{-}      & \multicolumn{1}{l|}{-}      & -     & -        & \multicolumn{1}{l|}{-}        & \multicolumn{1}{l|}{79.11} & 64.19 & -       \\ \hline\hline
    U-VB $\dagger$    & \multicolumn{1}{l|}{67.78}   & \multicolumn{1}{l|}{84.15}  & \multicolumn{1}{l|}{49.71}  & 57.89 & 56.53    & \multicolumn{1}{l|}{70.53}    & \multicolumn{1}{l|}{77.82} & 62.00 & 75.02   \\ \hline
    WFH & \multicolumn{1}{l|}{\textbf{68.41}}   & \multicolumn{1}{l|}{\textbf{84.82}}  & \multicolumn{1}{l|}{\textbf{50.50}}  & \textbf{58.46} & \textbf{58.39}    & \multicolumn{1}{l|}{\textbf{71.56}}    & \multicolumn{1}{l|}{\textbf{79.06}} & \textbf{62.73} & \textbf{75.91}   \\ \hline
    \end{tabular}}
    \label{table:vqa_rec_ve_results}
    \end{table*}
\vspace{-15pt}
\subsubsection{VQA, REC, and VE Tasks}\label{sec:vqa_rec_ve_results}
Table~\ref{table:vqa_rec_ve_results} mainly compares our proposed WFH models and U-VB. 
SOTA referred in the table represents the state-of-the-art task-specific models which do not follow the same "pre-trained and fine-tuned" paradigm. We refer SOTA as MCAN \cite{yu2019deep} on VQAv2, NSM \cite{hudson2019learning} on GQA, MAttNet \cite{yu2018mattnet} on RefCOCO+, and EVE-Image \cite{xie2019visual} on SNLI-VE as suggested in \cite{chen2019uniter}. Please note that since our model, along with U-VB, aims at being generic and versatile for different V-L downstream tasks, direct comparisons with those task-specific models are not the primary focus of this work. Instead, we compare against U-VB (U-VB $\dagger$) pre-trained in the VOLTA environment, while referring the readers to the reported results from the original U-VB work \cite{li-etal-2021-unsupervised}.

The proposed WFH consistently outperforms U-VB across all four tasks. Interestingly, comparing with VisualBERT (VB \cite{bugliarello2020multimodal}), which is pre-trained with text-image pairs, both WFH and U-VB achieve competitive results on VQAv2 test-dev split and SNLI-VE, while WFH offers much clear improvements on GQA (+1.81 points on test-dev split) and on RefCOCO+ (+1.86 points on test split).
    \begin{figure*}[ht!]
        \centering
        \begin{subfigure}[t]{.67\textwidth}
        \centering
        \includegraphics[width=1.\linewidth]{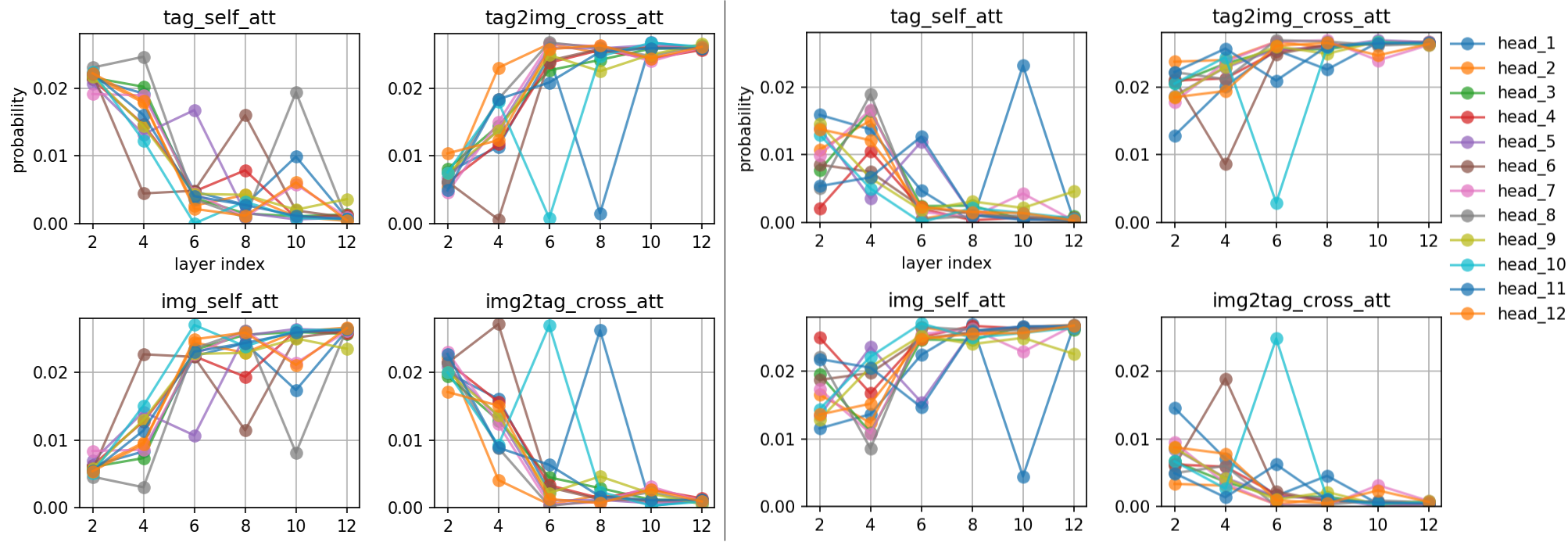}
        \caption{Left: U-VB. Right: The proposed WFH model.}
        \label{fig:atts}
        \end{subfigure}\hfill%
        \begin{subfigure}[t]{.33\textwidth}
        \centering
        \includegraphics[width=1.\linewidth,height=0.75\linewidth]{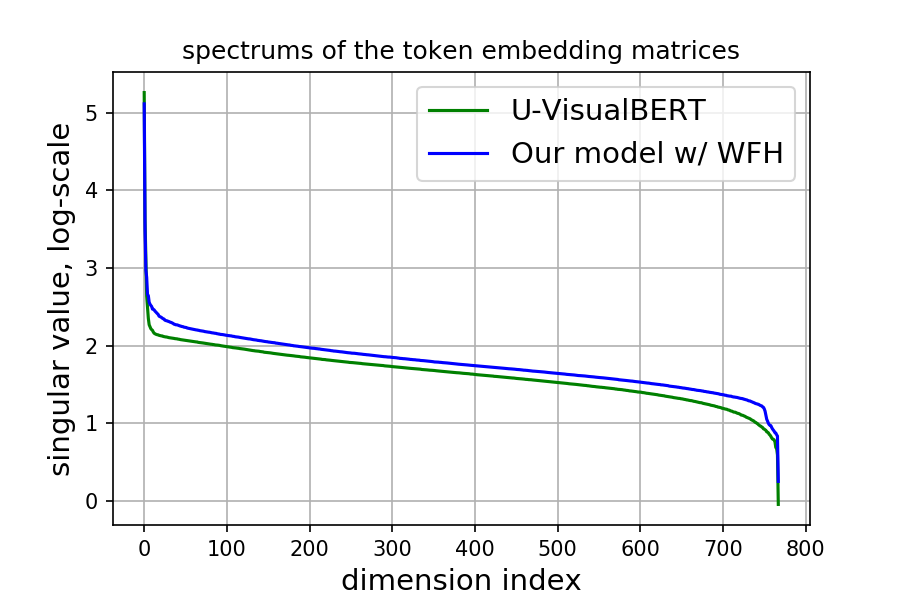}
        \caption{Spectrums from two pre-trained models.}
        \label{fig:spectrums}
        \end{subfigure}
        \caption{Analyses on W-VLP models. (a) Probabilities of attention heads in selected Transformer layers. (b) Spectrums, i.e. the ordered singular values ordered by magnitude, of the text token embedding matrices.}
        \label{fig:att_spec}
    \end{figure*}

\vspace{-12pt}
\subsubsection{What Can Attention Probabilities Tell Us?}
    In Figure~\ref{fig:atts} we dive into the probability distribution of each attention head learned over the $M=12$ Transformer layers across models. Both models exhibit similar patterns -- the attention probabilities on vision (\text{"img\_self\_att"} and \text{"tag2img\_cross\_att"}) keep increasing while those on language from tags (\text{"tag\_self\_att"} and \text{"img2tag\_cross\_att"}) decrease. This suggests that both models gradually find alignments across modalities by adapting the textual domain to the visual domain. More intriguingly, our model demonstrates much higher language-to-vision cross-attention probabilities (\text{"tag2img\_cross\_att"}), already from the early layers. This indicates that our model could benefit from the earlier cross-domain alignments, which are shown to be beneficial for a VLP model, supporting the similar finding in \cite{li2021align}. Thereby, we would like to emphasize that although the proposed WFH method is simple, it leads to fundamental changes in how two modalities behave and results in better transferability.
\subsubsection{What Can Spectral Analysis Tell Us?}
    We also compare the spectra of the word token embedding matrices in Figure~\ref{fig:spectrums}, i.e. the weight matrix involves in $T_{BERT}(\cdot)$ in Eqs.~(\ref{eq:token_transform}) and (\ref{eq:token_transform2}), of U-VB and our WFH model. Our model's spectrum decaying slower indicates that the word token embeddings our model generates are likely more expressive \cite{wang2019improving}. As such, the attention layers could be exposed with more diverse visual-textual embeddings from which they learn the alignment across the domains. This is especially crucial as the lack of paired V-L information is what the weakly-supervised model has to battle against. 

\vspace{-0.3cm}
\subsubsection{WFH with Different Configurations}
    Please refer to the SM for the studies, which analyze (1) how different ways of utilizing attribute tokens and (2) differently configured WFHs, e.g. with varying visual dictionary sizes, affect the downstream tasks.
\vspace{-0.1cm}
\subsection{Qualitative Studies on XMR}
    Please refer to the SM The studies are conducted on Flickr30K through the XMR tasks on which we compare how capable the considered models are in terms of aligning attributes, entities, and activities etc., across V-L domains.
\vspace{-0.2cm}
\section{Conclusion}\label{sec:conclusion}
    We proposed a novel W-VLP model that amends the lack of supervision from V-L pairs with a cross-domain hallucinator, dubbed as WFH, which generates bridging representations to interact with the textual modality. 
    
    Empirically, we found that the WFH model (1) learns more expressive word token embeddings, and (2) exhibits cross-domain alignments in the earlier Transformer layers.
    In retrieval tasks, it made consistent improvements, especially on the challenging cross-dataset generalization tests where it achieved at least 14.5\% gains in R@1 over U-VB. 
    The effectiveness of WFH was further confirmed in other V-L downstream tasks.
    
    Next, we will study how much the WFH models generalize to the downstream tasks given varying amounts of supervision, e.g. the number of different tags used. In addition, the current W-VLP models cannot be considered fully unpaired because they rely on a trained object detector, which is trained on image and the class labels -- a form of paired image-text data. We will explore ways to address this limitation to facilitate unpaired vision-language pre-training.
    
\section*{Acknowledgement}
This work has been supported by the Academy of Finland in projects 317388, 329268 and 345791. Special thanks to Aalto Science IT and CSC -- IT Center for Science, Finland for providing computing resources.
{\small
\bibliographystyle{ieee_fullname}
\bibliography{egbib}
}

\end{document}